# Transparent Adaptive Learning via Data-Centric Multimodal Explainable AI

MARYAM MOSLEH, Newcastle University, United Kingdom

MARIE DEVLIN, Newcastle University, United Kingdom

ELLIS SOLAIMAN, Newcastle University, United Kingdom

**Abstract**. Artificial intelligence-driven adaptive learning systems are reshaping education through data-driven adaptation of learning experiences. Yet many of these systems lack transparency, offering limited insight into how decisions are made. Most explainable AI (XAI) techniques focus on technical outputs but neglect user roles and comprehension. This paper proposes a hybrid framework that integrates traditional XAI techniques with generative AI models and user personalisation to generate multimodal, personalised explanations tailored to user needs. We redefine explainability as a dynamic communication process tailored to user roles and learning goals. We outline the framework's design, key XAI limitations in education, and research directions on accuracy, fairness, and personalisation. Our aim is to move towards explainable AI that enhances transparency while supporting user-centred experiences.

## 1. Introduction

AI and personalised learning have driven the development of adaptive learning systems. While these systems have made significant strides in tailoring content to learners, the explanations behind AI-driven decisions remain opaque and generic. [18]. Adaptive learning systems adapt to learners' performance and preferences by constantly tailoring the education style and tasks based on insights derived from learner engagement with the learning material [16, 20]. Data is collected and analysed through various AI and data analytics tools, including machine learning, Bayesian networks, neural networks, and educational data mining [10]. Despite personalised content delivery, these systems often lack transparency [6]. The rationale behind content selection and learner assessment remains unclear, creating a 'black box' effect.

This challenge can undermine trust among learners and educators and negatively impact their engagement when they're unsure of the system's validity and relevance to their learning journey [25]. While current XAI techniques use textual and visual explanations, most adaptive learning systems mainly focus on only providing non-personalised text explanations and rarely use personalised visual aids [18]. This limitation can negatively impact the effectiveness of AI decision explanations for learners with diverse preferences and needs [27]. While visual explanation techniques such as heatmaps and saliency maps are some of the most used visualisation methods in XAI, the general use of visual explanations in education remains limited [29]. In response to these limitations, it is essential to integrate personalised and context-specific XAI approaches into AI-driven adaptive learning systems. This paper presents a hybrid, user centric explainability framework that uses traditional XAI methods and generative AI to create personalised, multimodal explanations tailored to the needs of diverse educational stakeholders.

The key contributions of this paper are the following:
1) A review of current XAI limitations in education.
2) A novel hybrid framework combining traditional and generative XAI.
3) A conceptual pipeline for user-specific explanation delivery.
4) A roadmap for operationalising personalised explainability.

The remainder of this paper is organised as follows: **Section 2** reviews current XAI techniques and their educational limitations. **Section 3** introduces our proposed hybrid framework, supported by a conceptual architecture. **Section 4** outlines open challenges and research directions. **Section 5** discusses the potential impact, and **Section 6** concludes with future work.

## 2. Limitations of Current XAI in Education

Despite XAI techniques' progress in recent years, their integration in adaptive learning systems has been limited. Unfortunately, most of these XAI techniques, including SHAP, LIME, and counterfactual explanations, rely on algorithmic interpretability, hence static and non-personalised outputs. They follow a "one-size-fits-all" approach that provides explanations that don't cater to the preferences of different users, making AI transparency less inclusive [3].

Moreover, these techniques rarely consider the distinct needs of various user groups, such as students, instructors, administrators, and other stakeholders [18]. Each may require different levels of depth and presentation formats in AI explanations. Failing to address these diverse requirements can limit transparency and reduce trust in adaptive learning systems.

### 2.1. Common XAI Techniques and Their Educational Constraints

This section reviews widely used XAI techniques and highlights their educational limitations, focusing on how each method supports (or fails to support) personalisation, clarity, and multimodal delivery.

**SHAP** (SHapley Additive exPlanations) is an explainability method based on Shapley values from cooperative game theory. It assigns the importance of each feature in a machine learning model's prediction, indicating its impact on the model's final decision, where it provides consistent and mathematically validated explanations [21] . SHAP is widely adopted in financial institutions to interpret complex machine learning models, especially in credit scoring helping lenders make well-informed decisions [9]. However, learners without specialised knowledge may struggle to understand such outputs.

**LIME (Local Interpretable Model-Agnostic Explanations)** uses an ap-proximation technique that explains AI decisions by generating a local interpretable model around a given instance. It alters the input data to observe the shift in predictions, making it clearer to understand the process of AI decisions [22]. However, methods like SHAP and LIME often assume technical understanding and produce static explanations ill-suited for dynamic, user-focused educational settings. Also, the outputs tend to be sensitive to small changes in data, which introduces inconsistent interpretations. Moreover, LIME's effectiveness is further restricted due to the difficulty of handling high-dimensional, sequential or multimodal educational datasets, which limits its usefulness for complex learner paths.

By providing "what if" scenarios, **Counterfactual explanations** will present how slight alterations in the input data can affect the final outcome, helping learners understand the boundaries and the basis on which the model decisions are based [11]. Similarly, counterfactual explanations may suggest unrealistic interventions unless grounded in educational logic. However, they can propose unfeasible or educationally unsuitable interventions without careful adherence to pedagogical logic and guidance by education logic.

While SHAP, LIME, and Counterfactual explanations are well-known XAI techniques, a broader landscape of methods may be particularly valuable in educational settings.
Approaches like: **Anchor** intuitively generates high-precision decision rules [26]. Although these rules enhance interpretability beyond what feature attribution techniques offer, Anchor fails to capture the learning process's actual complexity and non-linear nature. **Surrogate Decision Trees** mimic complex models with simpler, interpretable decision trees [12]. Despite their clarity, they may overlook subtle model behaviours and interactions that are particularly important to consider in adaptive learning systems. For example, learners' prior knowledge, engagement level, and response time can all impact the system's content suggestions. **Gradient-based visual explanation methods** (e.g., saliency maps or Grad-CAM) [28] can offer clear insights into model reasoning by highlighting the areas in the input that significantly influence the model's decisions.

While these techniques perform well in visual tasks (e.g., image-based emotion recognition), they're less effective with textual education data. To develop effective and inclusive AI explainability for adaptive learning, we must investigate how these techniques can be tailored to varied educational needs and contexts. This exploration involves evaluating each method's interpretability, computational feasibility, and user preference, ensuring that the final system supports a wide range of learner profiles.

## 2.2. Common Comparison of XAI Techniques in Education

Table 1. Comparison of XAI Techniques in Educational Contexts.

| Method | Strengths | Weaknesses in Education | Level of Personalisation | Visual/Text Capabilities |
| --- | --- | --- | --- | --- |
| SHAP | Strong theoretical foundation | Too technical for most users | High- It can be tailored with learner data | Graphs and textual explanations |
| LIME | Straightforward and widely compatible | Struggles with complex or mixed formats | Low – static explanations | Basic text and plots |
| Counterfactuals | Learner based improvement cues | May suggest unrealistic actions | High – tailored via learner input | Text-based, with visual option |
| Anchors | Generates accurate understandable explanations | May oversimplify user needs | Medium - customisable, but not adaptive | Primarily text-based explanations |
| Surrogate Decision Trees | Clear via if-then-else rules | Poor fit for complex, non-linear systems | Medium – limited personal relevance | Visual trees and textual explanations |
| Gradient Visualisation Methods | Real-time insight into key factors | Limited causal insight | Low – not learner-specific | Visual outputs (e.g., heatmaps) |

## 2.3. Common XAI Techniques and Their Educational Summary of Gaps

For XAI techniques to be effective in educational settings, they require more than generic feature attribution; they must offer context-specific, learner-friendly and educationally meaningful explanations.

However, current existing methods generally fail to address the following criteria:
- **Contextual interpretability**: Explanations must be personalised based on different users' backgrounds and learning paths [17].
- **Multimodal integration**: The need to handle different data types, both visual and textual.
- **Transparency**: Enabling users to understand and influence the reasoning behind AI's decisions. Despite their potential, current XAI techniques rarely address the varied interpretability needs of educational stakeholders. Their limitations in delivering adaptive, multimodal explanations under-line the need for a new framework that places user context at its core.

### 2.4. Common Adaptive Learning Systems and Their Educational Summary of Gaps

Table 2. Comparison of Adaptive Learning Systems with the proposed framework.

| Feature/Aspect | AutoTutor | GnuTutor | ALEKS | Knewton | Our Framework |
|---|---|---|---|---|---|
| Focus | STEM tutoring with conversation | Replicating AutoTutor | Mastery learning in STEM | Adaptive content | Personalized explainable AI |
| Explainability | Moderate: Basic emotion and scripts | Low: Fixed, scripted responses | Low: Outcome-only focus | Low: black-box models | High: Adaptive, multimodal feedback |
| Personalization | Minimal: No persistent learner models | Minimal: (template-driven) | Medium: Adaptive knowledge model | High (data-driven predictions) | High (context-aware, preference-based) |
| Collected Data Types | Typed responses, timing, emotion (limited) | Typed responses, interaction logs. | Answers to concept-specific problems | Usage and performance data | Interaction data, user preferences, role context |
| Target Users | Learners only | Educators, Learners, researchers, and developers | K–12 and higher education | Higher education | Student, teachers, admins |

As shown in **Table 2,** AutoTutor offers real-time adaptation through Latent Se-mantic Analysis (LSA), cognitive state tracking and emotion detection, it is ultimately constrained by a finite-state model built on pre-scripted responses and limited set of instructional moves (e.g., prompts, hints, affirmations), limiting adaptation to scripted sequences and shallow interpretation of context.

The selected instructional moves are only guided by immediate user input and surface-level emotional cues. There is no generative feedback, adaptation to learning profiles, or tailoring of teaching strategies based on users' preferences [23].

GnuTutor is an open-source implementation that replicates AutoTutor's functional elements, such as LSA-based semantic analysis, speech act classification, mixed-initiative dialogue, and animated agents, while re-moving licensing barriers and providing a simpler way of deployment [24]. Although Gnututor's prolog (a logic-based language, used in AI and dialogue systems [24]) based architecture provides a more accessible codebase, like AutoTutor, it is still bound to finite-state and script-driven interactions, lacking the flexibility to tailor instructions based on the user's ongoing needs and preferences.

As a result, personalisation remains limited and reactive (triggered by immediate learner input, including behavioural or emotional cues). It lacks a mechanism for continuous user modelling, and thus cannot adjust instructional formats, such as depth, based on prior learner interactions or preferences [5]. While AutoTutor and GnuTutor can excel in delivering structured approaches to tutoring, they still share the same limitations of static personalisation, script-bound interaction, a lack of generative adaptability to user's preferences and needs, and the lack of transparent reasoning processes as they don't provide explanations or personalisation features that our proposed framework does. By integrating XAI, generative models and real-time user data modelling, our system supports broader functionality across different user roles.

While Autotutor and GnuTutor rely on structured pedagogical dialogue and pre-scripted responses, Knewton adopts a data-centric approach that integrates psycho-metric profiling, collecting and analysing users' data using AI to estimate their skills, preferences, needs, and knowledge [30]. Furthermore, it utilises content graph alignment, where it structures the learning material in the form of knowledge graph that incudes aligning interconnected concepts with learner's current level of understanding and performance, guiding the system into the next best step and concept based on what the learner has improved in and what's connected to it [2, 30]. Despite offering performance metrics and visual dashboards, the underlying logic behind the adaptive decisions remains inaccessible, raising concerns regarding the reliability of the system's automated interventions [14].

Although some adaptive learning systems provide explanations for AI decisions, the AI models used often depend on pre-set rules and parameters, resulting in static and non-personalised explanations. This inflexibility could negatively impact the user's experience as the explanations do not account for their different needs and preferences [27]. For example, the instructor's manual of the Assessment and Learning in Knowledge Spaces (ALEKS) system includes "explanations and algorithmically generated practice problems", stating that the explanations of the material and AI decisions are not tailored to the user's needs and preferences [1]. Instead, ALEKS provides standardised explanations for all users, following a "one-size-fits-all" approach.

The study by Conati et al. [7] on tutoring systems illustrates how personalised explanations of AI decisions can improve learners' trust and engagement with these systems, emphasising the importance of designing future learning systems that don't only focus on delivering accurate content but also provide different explanation techniques that adapt to learners' unique profiles, hence adopting a human-centric perspective. This aligns with our proposed framework's aim to provide personalised, multimodal explanations, promote transparency with AI decisions, and improve users' trust, thus closing the gap between the AI decision-making process and users' understanding.

## 3. Proposed Framework

Despite existing research on personalisation, current XAI techniques fail to deliver meaningful, user-specific explanations. This creates a disconnect between adaptive systems' potential and user experience. Although adaptive learning systems can ad-just learning content based on users' progress, the provided explanations are often static and generic, making it difficult for learners to understand the reasoning behind AI recommendations. The rationale for combining generative AI with traditional XAI is grounded in the challenge that most raw explanations from SHAP, LIME, or counterfactual methods are either overly technical or not adapted to learner roles. Generative AI models have shown promise in translating structured data into natural language that matches user comprehension levels (e.g., OpenAI's GPT-4o use in education as shown in Kim et al.'s (2024) study [19], where it was used to tailor scientific information to individual learner profiles and shown its effectiveness in improving user's understanding). We extend this idea by using generative models as a translation layer for XAI outputs. We propose a hybrid explainability framework to address this gap, which will generate adaptive, multimodal explanations tailored to user roles and preferences.

### 3.1. Overview of the Hybrid Framework:

The framework design includes four main stages:

- **Data Collection and Learner Profiles:** We begin by analysing existing XAI methods used in adaptive learning systems to identify their current limitations. In parallel, we will categorise educational stakeholders (such as students, teachers, module leaders, and administrators), and determine their specific explainability needs through interviews, focus groups, and surveys. This process enables the creation of dynamic learner profiles that accurately reflect user knowledge, goals, and contexts.
- **AI Decision Engine and XAI Layer:** Learner data will be processed by an adaptive learning system using models such as Bayesian Knowledge Tracing (BKT) [4] to tailor educational content. The decisions generated by these models will be interpreted via a dedicated XAI layer, which selects and applies the most suitable explainability method according to each user's profile and preferences.
- **User-friendly explanations through Generative AI:** Generative AI will convert technical XAI outputs into accessible, conversational explanations. For example, rather than presenting a technical explanation like: "SHAP value of -0.3 for concept node algebraic expressions" the system would generate a more user-friendly message such as: "We noticed that you spent extra time solving the last two algebraic expression problems, so we are offering additional practice to improve your understanding."
- **Personalisation**: Explanation delivery will be tailored to each user type and context:
    - **Students** will receive simple, motivational explanations in both text and visual.
    - **Teachers** will access detailed dashboards showing student progress, knowledge gaps, and performance-enhancing suggestions.
    - **Administrators** will be provided with high-level system trend summaries and user engagement reports.

This framework therefore aims to produce explainable AI and also deliver explainable-to-the-user AI. Explanations are personalised in language style, delivery format, and depth, aligning with individual user roles, preferences, and cognitive needs.

### 3.2. Conceptual Pipeline Diagram Description

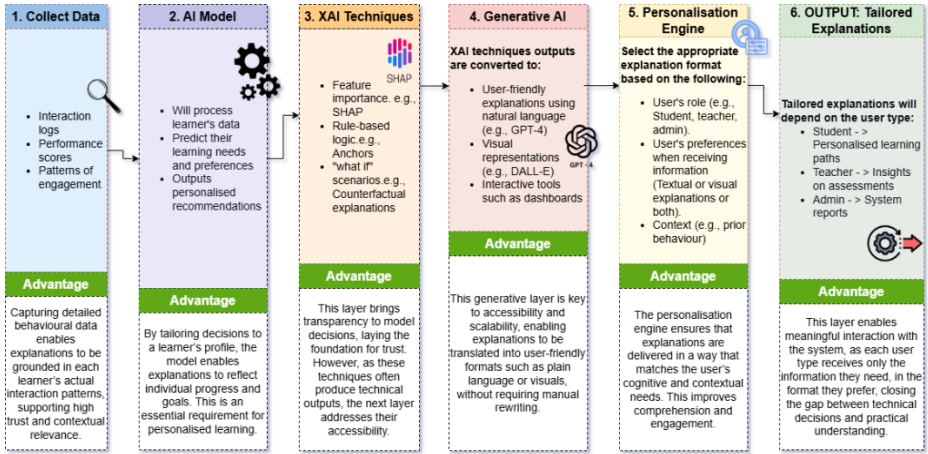

Fig. 1. Conceptual Architecture of the Proposed Hybrid Explainability Framework

Designed for adaptive learning environments, the suggested framework introduces a layered conceptual pipeline to support delivering personalised, multimodal explanations of AI decisions in these systems. Figure 1 presents the overall architecture of our proposed hybrid explainability framework. It consists of six layers: data collection, AI decision-making, XAI explanation generation, generative AI translation, a personalisation engine, and delivery of tailored outputs. This layered approach ensures that explanations are technically accurate, context-aware, user-friendly, and aligned with individual learning roles and needs.

Each layer is described in detail below:

- **Layer 1:** Collect Data: The system collects continuous learner data (performance, engagement) to inform decisions. This provides the AI model with the rich data needed to understand each user's different learning paths.

- **Layer 2:** AI Model: The AI model then uses the collected data to generate personalised recommendations and predictions, such as suitable learning materials and potential areas of misunderstanding. However, the decision's lack of explanation still challenges learners and educators to understand or trust it fully.

- **Layer 3:** XAI Techniques: To tackle this problem and clarify how AI decisions are made, this framework will integrate different XAI techniques that will help un-cover the reasoning behind AI decisions by emphasising the key features affecting its outputs.

- **Layer 4:** Generative AI: Though technically valid, XAI outputs are often com-plex. Generative AI simplifies them into accessible text or visuals.

- **Layer 5:** Personalisation Engine: Is the decision-making layer. By analysing user roles, preferences, and interaction context, it will dynamically select the most appropriate explanation method, format, and depth for each individual. This layer includes: (1) user profile identification (e.g., student, teacher, admin); (2) contextual analysis to align content and explanations with user goals; (3) format selection based on profile and context; and (4) a feedback loop that updates profiles from ongoing engagement.

- **Layer 6:** OUTPUT: Tailored Explanations: The final personalised explanations are delivered at this layer with a suitable depth and format based on the user's requirements and preferences. For example, learners will receive adaptive feedback tied to their progress, instructors or teachers will receive specific insights to inform their teaching decisions, and administrators will receive high-level performance monitoring reports.

### 3.3. Illustrative Examples of Personalised Explanations

To illustrate how our framework operates, we present hypothetical examples de-signed to reflect the needs of different user types:

- **Student**: *"You've done well on basic recursion problems but got stuck when it in-volved trees. That's common — tree problems are harder because they involve multiple recursive calls. Here's a step-by-step example to help you practice breaking it down."*

- **Teacher**: *"Several students in your class struggled with dynamic programming this week. Most made the same mistake: not storing previous results. A short group re-view using visual aids on memorization might help reinforce the core idea."*

- **Administrator**: *"Data shows that first-year students in the online program are spending significantly more time on introductory algorithm modules compared to their in-person peers. This could indicate a need for additional scaffolding or better pacing online."*

### 4. Key Challenges and Open Questions

While the vision for Human-Centric, Multimodal Explainable AI in adaptive learning systems holds immense potential, it also raises foundational questions and risks. Instead of offering definite answers, this section seeks to highlight these questions and risks that could guide cross-disciplinary investigation.

### 4.1. Accurate, Faithful and Personalised Explanations

**Question**: *How can personalised explanations remain accurate and faithful to the model?*

Personalised explanations are intended to adapt to different user roles and cognitive needs; however, they must also accurately reflect the model's underlying logic and reasoning. The challenge lies in preserving the fidelity of AI decisions while adapting their presentation to suit learners, teachers, or administrators. The accuracy and relevance of these personalised outputs may not always align with established standards of explainability, such as user transparency, as highlighted in prior work, including EU AI regulation [8]. This raises concerns about whether tailoring explanation for each user might inadvertently introduce bias, misrepresentation, or inconsistency. Techniques that validate the educational and logical integrity of personalised outputs will be essential to ensure both trust and utility. To address this within our research, the system will include validation mechanisms ensuring that generated explanations faithfully reflect the underlying model's reasoning, verified through expert review and alignment with model outputs.

### 4.2. Understandability, Accuracy, and Fairness

**Question**: *What trade-offs exist between understandability, accuracy, and fair-ness?*

Even when fidelity to the model is preserved, adaptive learning systems must still address the inherent tension between explanation clarity and technical detail. Over simplifying explanations to improve accessibility may reduce their usefulness or lead to misunderstandings, while maintaining full technical accuracy may make them incomprehensible to many users. In educational contexts, these trade-offs can influence learner development, engagement, and trust in AI systems. Prior work by Holstein et al. [13] also stresses the importance of fairness and transparency in AI systems. Solutions require testing with users to balance clarity, accuracy, and fairness. In our work, this trade-off will be empirically studied in user trials, where multiple explanation types (e.g., simple vs. detailed) will be compared across different learner pro-files to identify optimal balances.

### 4.3. Generative AI in education

**Question**: *Can generative AI be reliably used to explain critical decisions in education?*

Recent advances in large language models (such as ChatGPT) have introduced powerful new capabilities for delivering conversational and adaptable explanations and supporting interactive learning. However, these models still raise serious concerns regarding the accuracy of the information provided, including the reasoning behind the model's decisions. Generative AI may generate biased, irrelevant, or inaccurate results. Appropriate control mechanisms will be implemented to address this, and the model will be fine-tuned using a diverse dataset, followed by iterative testing and refinement. Furthermore, multiple validation layers will be required to evaluate the reliability of the generated explanations. This includes evaluations from experts in the education sector to verify the pedagogical accuracy. At the same time, students' feedback that will be gathered through user studies in low-risk and controlled settings (e.g., formative assessments) can assess the interpretability of the generated explanations and their usefulness. To ensure reliable generative outputs, models will be fine-tuned on educational datasets and constrained through templates aligned with AI decisions, combined with human validation.

### 4.4. Adapting to user explanation preferences

**Question**: *How should explanation preferences be modelled and updated for different users?*

Every user has unique preferences, needs, and requirements; some learners may prefer visual explanations, while others favour textual explanations or detailed, structure guidance. These preferences may change over time as users become more skilled, presenting the need for dynamic adaptation within learning systems. User preferences will be initially captured through onboarding surveys and refined through interaction data (e.g., skipping visuals), with dynamic updates via a feedback loop.

## 5. Potential Impact

The proposed framework has the potential to significantly enhance the interaction experience between learners and education technologies. Delivering personalised and user-friendly explanations in real time helps address the critical challenges of AI-driven adaptive learning systems and educational AI by ensuring clarity and fairness.

**Potential Impact on Students:**
- Enhance system's transparency, hence improving users' trust in AI decisions and engagement with learning systems.
- Promoting metacognitive awareness, encouraging users to self-monitor and evaluate their progress on their learning journey [15].
- Users' roles will change from passive consumers to collaborators in their learning process.

**Potential Impact on Teachers:**
- Gain deep and transparent reports on student engagement and learning progress.
- Traditional performance metrics will be replaced with personalised explanations, allowing teachers to see the adaptive logic driving AI decisions.
- Enhance the interaction between users and AI, supporting a data-driven and responsive teaching practice.

**Potential Impact on Institutions:**
- This framework will ensure the main criteria of responsible AI, accountability, fairness, transparency, and adherence to institutional and legal guidelines, as institutions can align algorithmic behaviour with their ethical standards, domain-specific benchmarks, and accreditation bodies' expectations.
- It can help institutions achieve their educational goals and enhance student development through early intervention. It ensures that the decisions and recommendations it provides align with the set goals, such as improving learning outcomes and supporting student reflection and confidence building ahead of summative assessments.
- Explainable AI decisions support external validation, where parents can easily understand their child's learning path and progress, and policymakers can easily evaluate the system.

**Potential Impact on Other Domains**: Healthcare, Finance and Social Platforms: Although education remains the primary focus of this paper, the core principles of this framework can be applied to other high-stakes sectors where trust and human-AI interaction are critical parts of them. In the healthcare sector, personalised explanation could aid patients' understanding of treatments.

In finance, personalised rationales for decisions that depend on algorithms, such as credit scoring, could support users to make an informed decision. In social platforms, adaptive explanations can provide the clarity needed to understand certain content choices, including filtering or prioritising.

## 6. Conclusion

We presented a vision for hybrid explainability in adaptive learning systems that integrates user modelling, traditional XAI, and generative AI. This human-centric approach aims to enhance transparency, comprehension, and trust. In the next phase, we will conduct an extensive literature review to discover and evaluate the current AI-driven adaptive learning systems and identify their limitations in explaining AI decisions. This will then be followed by conducting user studies within Newcastle University's School of Computing to assess the impact of explanation types on trust and engagement. Methods such as surveys, focus groups, and semi-structured interviews will be employed to understand what each group considers a 'meaningful' explanation and how they prefer to receive it. Findings from this phase will guide the selection or development of XAI techniques and the generation of personalised explanations. Next is systematically evaluating a range of XAI and how Generative AI can complement these methods. The system will be tested with Newcastle University School of Computing students (e.g., Students from a specific module within the School of Computing) and educators. Data will be collected via workshops, inter-views, and questionnaires to assess trust, understanding, and engagement. Finally, a controlled study will be conducted with multiple experimental groups, each exposed to different forms of Al explanations (e.g., numeric-based, rule-based, text-based, visual, and hybrid). Participants' trust, understanding, and engagement levels will be measured. We will also analyse whether certain user groups (e.g., novices vs. advanced learners, teachers vs. admins) prefer specific explanation types. This paper outlines a forward-looking vision for hybrid explainability in adaptive learning. Future work will implement and evaluate this framework through participatory design with learners and educators.